# Vision-Based Environmental Perception for Autonomous Driving

Fei Liu, Zihao Lu, Xianke Lin*

Department of Automotive and Mechatronics Engineering

Ontario Tech University

Oshawa, canada

*Corresponding author: xiankelin@ieee.org

## Abstract

Visual perception plays an important role in autonomous driving. One of the primary tasks is object detection and identification. Since the vision sensor is rich in color and texture information, it can quickly and accurately identify various road information. The commonly used technique is based on extracting and calculating various features of the image. The recent development of deep learning-based method has better reliability and processing speed and has a greater advantage in recognizing complex elements. For depth estimation, vision sensor is also used for ranging due to their small size and low cost. Monocular camera uses image data from a single viewpoint as input to estimate object depth. In contrast, stereo vision is based on parallax and matching feature points of different views, and the application of deep learning also further improves the accuracy. In addition, Simultaneous Location and Mapping (SLAM) can establish a model of the road environment, thus helping the vehicle perceive the surrounding environment and complete the tasks. In this paper, we introduce and compare various methods of object detection and identification, then explain the development of depth estimation and compare various methods based on monocular, stereo, and RDBG sensors, next review and compare various methods of SLAM, and finally summarize the current problems and present the future development trends of vision technologies.

Keywords： vision sensor, vision SLAM, environmental perception, Deep Learning, autonomous driving

## 0 Introduction

Environmental perception is one of the most important functions of autonomous driving. The performance of environmental perception, such as accuracy, robustness to light changes and shadow noise, and adaptability to complex road environments and bad weather, directly affect the performance of autonomous driving technology. The commonly used sensors in autonomous driving include ultrasonic radar, millimeter wave radar, LIDAR, vision sensors, etc. Although global position technology, such as GPS, BeiDou, GLONASS, etc., is relatively mature and capable of all-weather positioning, there are problems such as signal blocking or even loss, low update frequency, and positioning accuracy in environments such as urban buildings and tunnels. Odometer positioning has the advantages of fast update frequency and high short-term accuracy, but the cumulative error over the long term is large. Although LIDAR has high accuracy, there are several disadvantages, such as large size, high cost, and weather-dependent. Tesla and several companies, such as Mobileye, Apollo, and MAXIEYE, use vision sensors for environmental perception. The application of vision sensors in autonomous driving is based on cameras with advanced artificial intelligence algorithms that facilitate object detection and image processing to analyze obstacles and drivable areas, thus ensuring that the vehicle reaches its destination safely[1]. Visual images are extremely informative compared with other sensors, especially color images. They contain not only the distance information of the object but also the color, texture, and depth information, thus enabling simultaneous lane line detection, vehicle detection, pedestrian detection, traffic sign detection through signal detection, etc. Also, there is no interference between cameras on different vehicles. The vision sensor can also achieve simultaneous localization and map building (SLAM). The vision information is obtained from real-time camera images, providing information that does not depend on a priori knowledge and has a strong ability to adapt to the environment.

The main applications of vision-environmental perception in autonomous driving are object detection and identification, depth estimation, and SLAM. Vision sensors can be divided into three broad categories according to how the camera works: monocular, stereo, and RGB-D. The monocular camera has only one camera, and the stereo camera has multiple cameras. RGB-D is more complex and carries several different cameras that can read the distance of each pixel from the camera, in addition to being able to capture color images. In addition, the integration of vision sensors with machine learning, deep learning, and other artificial intelligence can achieve better detection results [2]. In this paper, we will discuss the following three aspects.
1) Vision-based object detection and identification, including traditional methods and methods based on deep learning;
2) Depth estimation based on monocular, Stereo, and RGBD and the application of deep learning;
3) Monocular SLAM, Stereo SLAM, and RGBD SLAM.

## 1 Object detection and identification

### 1.1 Traditional object detection and identification methods

In autonomous driving, identifying road elements such as roads, vehicles, and pedestrians and then making different decisions are the foundation for the safe driving of vehicles. The workflow of object detection and identification is shown in Figure 1. The image acquisition is made by cameras that take pictures of the surrounding environment around the vehicle body. Tesla [3] uses a combination of wide-angle, medium-focal length, and telephoto cameras. The wide-angle camera has a viewing angle of about 150° and is responsible for recognizing a large range of objects in the near area. The medium focal length camera has a view angle of about 50° and is responsible for recognizing lane lines, vehicles, pedestrians, traffic lights, and other information. The view angle of the long-focus camera is only about 35°, but the recognition distance can reach 200~250 m. It is used to recognize distant pedestrians, vehicles, road signs, and other information and collect road information more comprehensively through the combination of multiple cameras.

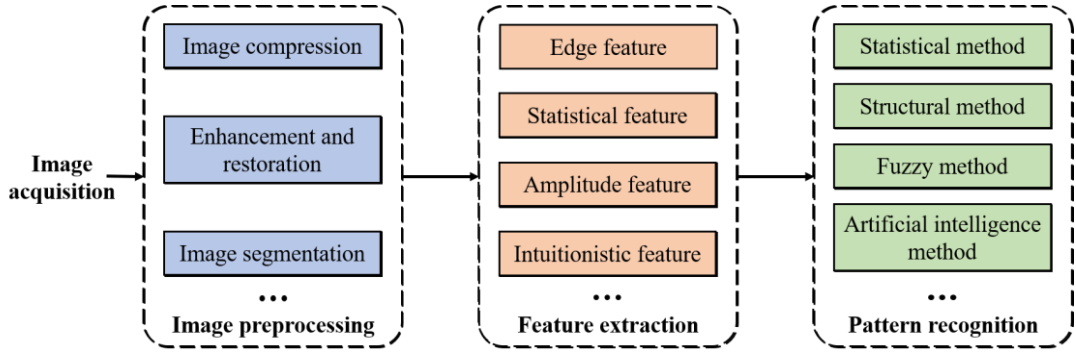

Figure 1 Object detection and identification process, including image acquisition, image preprocessing, image feature extraction, image pattern recognition, etc.

Image preprocessing eliminates irrelevant information from images, keeps useful information, enhances the detectability of relevant information, and simplifies data, thus improving the reliability of feature extraction, image segmentation, matching, and recognition. This process mainly includes image compression, image enhancement and recovery, image segmentation, etc.

（1）Image compression can reduce the processing time and the memory size required. Currently, image compression methods include discrete Fourier transform compression [4], discrete cosine transform compression [5], NTT (Number Theory Transformation) compression [6], neural network compression [7], wavelet transform compression [8, 9], and so on. Among them, wavelet transform is more widely used because of its high compression ratio, fast compression speed, and strong anti-interference capability. Furthermore, the grayscale image can compress the color image consisting of Red, Green, and Blue channels acquired by the vision sensor into a grayscale map that is represented by grayscale values only. In this way, the distribution and characteristics of color and brightness of the image can be fully reflected, and the processing time is reduced. The common methods for grayscale images include the fractional method, the maximum method, and the average method.

（2）Image enhancement and recovery are used to improve image quality, remove noise, and improve image clarity. Image enhancement techniques are mainly divided into the spatial domain method and the frequency domain method. The spatial domain method [10–14] is mainly used to directly compute pixel grayscale values in the null domain, such as image grayscale transform, histogram correction, image null domain smoothing and sharpening, pseudo-color processing, etc. The frequency domain method is used to compute transformation values in some transformation domains of the image, such as the Fourier transformation [15].

（3）Image segmentation divides an image into several specific regions with unique properties and then extracts the target. This method is a prerequisite for image recognition, and its performance directly affects the quality of image recognition. The main image segmentation methods are threshold segmentation, region segmentation, edge segmentation, and specific theoretical segmentation methods such as mathematical morphology-based, neural network-based, genetic algorithm based, etc. The comparison is shown in Figure 2.

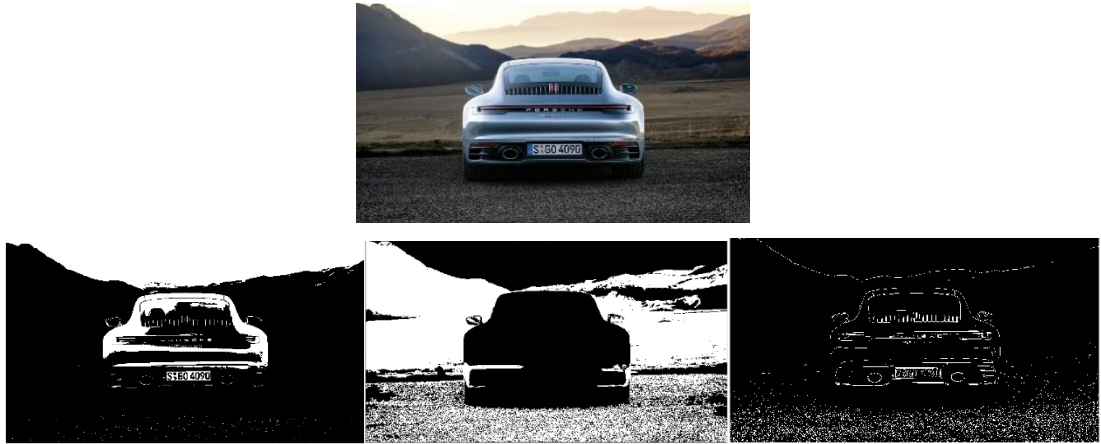

Figure 2 The top is the original image, and the bottom is the image after threshold, region, and edge segmentation, respectively

Table 1 Summary of image segmentation

| Image segmentation | References | Working Principle | Advantages | Disadvantages |
|---|---|---|---|---|
| Threshold segmentation method | [16–19] | Use colors, grayscale, contour, and other selection thresholds to separate the target object from the background | Quickly achieve image segmentation, often used in cases where there is a large difference between the target and background | Difficult to recognize the complex environment. Only for preliminary segmentation |
| Regional segmentation method | [20–23] | Use the whole image as a starting point to gradually exclude or merge similar pixels or select a pixel to merge pixels with similar features continuously | It has more advantages in pixel similarity and spatial adjacency, and has stronger robustness which significantly reduces the interference of noise | It is easy to cause over-segmentation of the image, which can combine edge detection to get better segmentation performance |
| Edge segmentation method | [24–27] | Detecting boundary points by using the characteristics of different gray values of different regions and a big change of gray values at the boundary, then connecting each boundary point for region segmentation | Fast and good detection of edges | Cannot obtain good area structure, and there is a contradiction between noise immunity and detection accuracy, where accuracy is improved at the expense of noise immunity |

　　It is necessary to extract the required features and calculate the feature values based on image segmentation in order to complete the identification of objects in images. The key to vehicle identification lies in quickly extracting features and achieving accurate matching. The main features are shown below.

（1）Edge Features

The detection operators for edge features include Canny operator [28], Roberts operator [29], Prewitt operator [30], Sobel operator [31], Laplacian operator [32], etc. The Canny operator has good resistance to noises and uses two different thresholds to detect strong and weak edges, respectively, which performs even better for detecting objects with blurred boundaries. The Roberts operator is better for images with steep low noise, but the extracted edges are coarse, so the edge positioning is not very accurate. The Sobel and Prewitt operators are better for images with noise and gradual changes in grayscale value and are more accurate for edge positioning. The Laplacian operator locates the step edge points in the image accurately but is very sensitive to noise and easily loses orientation information resulting in discontinuous detected edges.

（2）Appearance Features

Appearance features mainly include edges, contours, texture, dispersion, and topological characteristics of the image. [33] obtained the Gray-level co-occurrence matrix (GLCM) by computing the grayscale image and then calculated the partial eigenvalues of the matrix to represent the texture features of the image. [34][35] completed the terrain classification by combining the geometric classifier and color classifier. Real-time ground information can be obtained with continuously updating the 3D data of the terrain surface collected by the vehicle driving.

（3）Statistical Features

Statistical features mainly include histogram features, statistical features (such as mean, variance, energy, entropy, etc.), and statistical features describing pixel correlation (such as autocorrelation coefficient and covariance). [36] performed stereo matching by extracting the vehicle taillight center point from the left camera and right camera, and detected the vehicle ahead using directional gradient histogram features.

（4）Transformation Coefficient Features

It includes the Fourier transformation, Hough transformation, Wavelet transformation, Gabor transformation, Hadamard transformation, K-L transformation, etc. [37] used an improved Hough transformation to extract small line segments of the lane. It identifies lanes by clustering small line segments using a density-based clustering algorithm with noise and by curve fitting. The experimental results show that the detection is better than the linear algorithm and is robust to noise.

（5）Other Features

Images include pixel grayscale values, RGB, HSI, and spectral values. Researchers in [38] proposed a vehicle detection method based on color intensity separation, which uses intensity information to filter the Region of Interest(ROI) of light changes, shadows, and clutter background, and then detects vehicles based on the color intensity difference between vehicles and their surroundings. [39] converted RGB video frames captured by RGB-D to color gamut images, whose noise can be reduced or eliminated for each frame. This method distinguishes the color characteristics of vehicles more accurately and achieves tracking of vehicles.

Object identification is performed based on the extracted features, which compares the object of interest with existing known patterns to determine its category. Object identification methods can be divided into different categories based on the features used, for example, shape features, color features, texture features, etc. Based on the identification methods used, it can be divided into statistical object identification [40], structural object identification [41], fuzzy object identification [42], neural network object identification [43], etc.

Due to the complexity of the road environment, vehicles must rely not only on a single forward-facing camera but also on the surrounding view. Blind spots have led to many car accidents, so detecting pedestrians and vehicles in blind spots is critical. In 2006, [44] proposed the concept of a panoramic surround view. Surround-view cameras or panoramic video monitoring (AVM) systems, as shown in Figure 3, stitch together images from all directions of the car and identify road signs, curbs, and nearby vehicles, making it easy for drivers to look around, thereby reducing the number of car accidents.

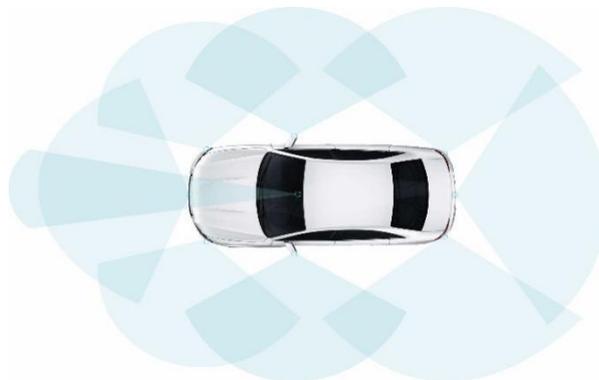

Figure 3 Surround-view cameras

In 2007, Nissan first used a surround-view camera. It stitched together a panoramic view from four cameras located on the front windshield, left and right-side mirrors, and above the license plate of the car. Fujitsu's Wrap Around View [45] provides a view that is updated in real-time. The driver can select the best view for different situations, including the "third-party" view and images of the vehicle itself and its surroundings. [46] further developed 3D AVM based on 2D AMV, which can expand the visual coverage, thus helping the driver to determine the collision distance with another vehicle on narrow roads and improving driving safety. Tesla's vision system adopts three cameras in the front of the car, one in the rear, and two in the side rear and side front, respectively, to achieve a total of eight cameras for accurate blind zone monitoring and target-ranging functions.

The main techniques used in Surround View Cameras include image correction, top view transformation, image matching, image fusion, etc. The images captured by the Fisheye Lens are distorted and therefore require calibration and correction. The most commonly used calibration method for Surround View Cameras is the Zhang Zhengyou calibration method [47], which yields stable results and is easy to use. The direct linear transformation method [48] uses the one-to-one correspondence between multiple 3D point coordinates and pixel coordinates to compute a linear system of equations to obtain the camera model. However, this method does not consider the nonlinear distortion of the camera lens, so the accuracy is not high. Therefore, the Tsai calibration method was proposed in [49]. The method first solves the linear relationship between pixel coordinates using the direct linear transformation method, takes this linear relationship as the initial value, and then combines the nonlinear distortion of the lens. It uses an optimization algorithm to optimize the internal and external parameters of the camera, which solves the low accuracy problem of the direct linear transformation method. In addition, there is the Scaramuzza calibration method [50], LADMAP method [51], etc. The frequently used fisheye correction algorithms can be divided into projection model-based correction methods and 2D and 3D space-based correction methods. Since the final rendering of the panoramic monitoring is a top view, the corrected map must first be transformed into a top view, mainly based on camera parameters and based on the projection matrix.

The image stitching methods include SIFT-based stitching methods [52] and SURF-based stitching methods [53] which are improved based on SIFT. The SIFT method first finds the extreme points, delete the points with small influence, and then uses the Hessian matrix to delete the edge points to calculate the description information of the feature points. Then, the feature points in each image are compared, and the points with similar features are considered as the same position for stitching. In [54], the quality of the panoramic view is enhanced by using six Fisheye Lens, improving the SIFT algorithm for image matching, and improving the fading-in and fading-out algorithm for better image fusion. SURF features use Haar wavelet response and image integration when calculating the descriptive information of feature points, thus collecting feature information quickly and improving the accuracy of feature point matching simultaneously.

**1.2 Deep Learning-based object detection and identification**

Compared with traditional object detection and identification, Deep Learning requires training based on a large dataset but brings better performance. Traditional object identification methods do feature extraction and classifier design separately and then combine them together. In contrast, deep learning has more powerful feature learning and feature representation capabilities by learning the database and mapping relationships to process the information captured by the camera into a vector space for recognition through neural networks. The object detection and identification model is shown in Figure 4. The "Backbone" in the figure refers to the convolutional neural network for feature extraction that has been pre-trained on a large dataset and has pre-trained parameters. The "Neck" represents some network layers used to collect feature maps in different stages. The "Head" represents the type and location of bounding boxes.

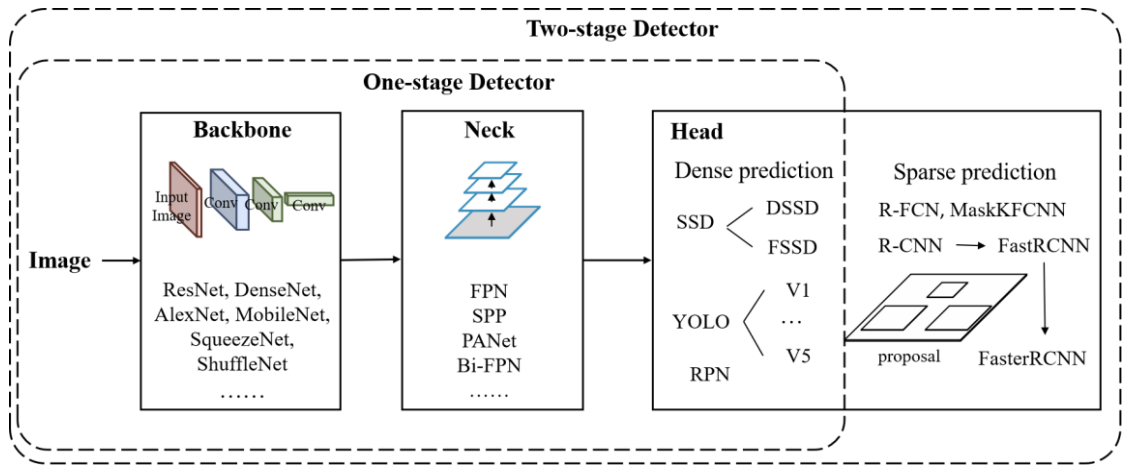

Figure 4 Object detection and identification module, mainly including Backbone, Neck, and Head

For the "Backbone", AlexNet proposed in [55] in 2012 was the first application of Deep Learning technology to large-scale image classification. Compared to other deep learning networks, FlexNet used five layers of convolutional layers and three layers of fully connected layers, the activation function replaced sigmoid with ReLU, and Dropout is used in the first two layers, i.e., randomly deactivating some cells to reduce the overfitting problem. The object identification error rate reaches 17%. A layer-by-layer greedy algorithm was proposed in [56], which overcomes the drawback that deep networks are difficult to train and reduces the data dimensionality with self-encoders. In 2014, [57] proposed a full convolutional neural network (FCN) to achieve pixel-level segmentation with a deep convolutional neural network approach, and [58] proposed a ZFNet to visualize the features learned by a CNN through deconvolution for object identification and achieved an error rate of 11.7%. In 2015, [59] proposed GoogLeNet using a Network in the network (NiN) based network, the Inception module of [60] is shown in Figure 5. By extracting information in parallel through convolutional and max-pooling layers of different sizes, the 1 × 1 convolutional layer can significantly reduce the number of parameters, decrease the model complexity, and reduce the identification error rate to 6.7%.

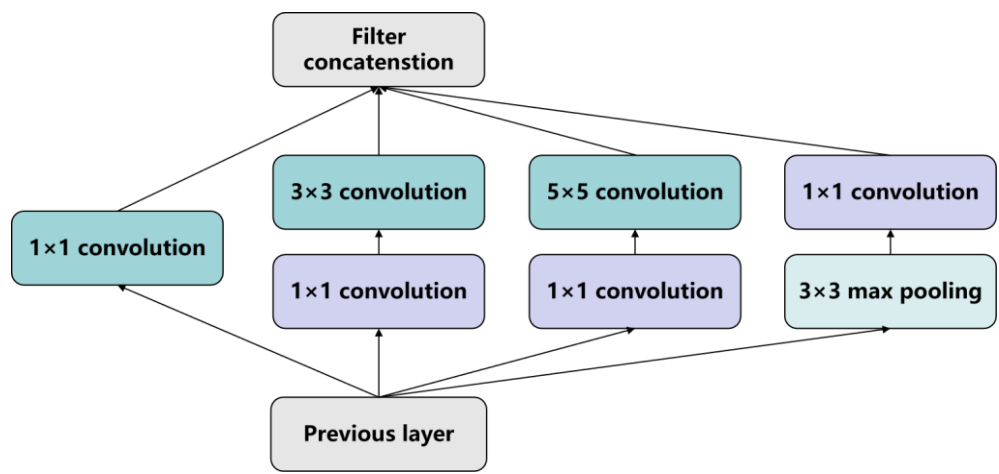

Figure 5 Inception module

In 2016, [61] proposed ResNet to solve the gradient vanishing problem during backpropagation. It improves the efficiency of information propagation by adding directly connected edges to the nonlinear convolutional layers to reduce the error rate to 3.6% and increases the network depth to hundreds of layers, so it can train deep networks without adding a classification network in the middle to provide additional gradients like GoogLeNet. [62] proposed DeconvNet, which adopted an inverse process opposite to the forward pass process and introduced deconvolution and anti-pooling layers to reduce the feature map to the original size and achieve the segmentation after the mirroring process. Google[63] proposed Deeplab V2 with an ASPP structure based on Deeplab V1, which can pool the original feature maps at different scales and then fuse the results of each scale to achieve better identification results. However, in traditional convolutional neural networks, when information about the input or gradient passes through many layers, it may disappear when it reaches the end of the network. In 2017, DenseNet [64] developed based on ResNet, introduced a direct connection between any two layers with the same elemental graph size, such that L connections were increased to L(L+1)/2 connections, as shown in Figure 6, where H(i) includes Batch Normalization (BN), ReLU, Pooling, and Convolution (Conv). The traditional feedforward architecture can be considered as an algorithm with a state that is passed layer by layer and therefore does not need to relearn redundant feature maps. One of the major advantages of DenseNet is that it improves the flow of information and gradients throughout the network, making it easier to train. Each layer can access the gradients directly from the loss function and the original input signal, which both drastically reduces the number of parameters in the network and alleviates the problem of gradient vanishing to some extent, and enhances the transmission of features of its structure. And in the same year, [65] proposed ResNeXt, which continues ResNet and Inception and adds residual connectivity. [66] proposed Deeplab V3, which optimized DeeplabV2. The Atrous Spatial Pyramid pooling (ASPP) structure in [63] introduced tandem multi-scale null convolution and used the Batch Normalization method to improve the segmentation accuracy, thus allowing better object identification.

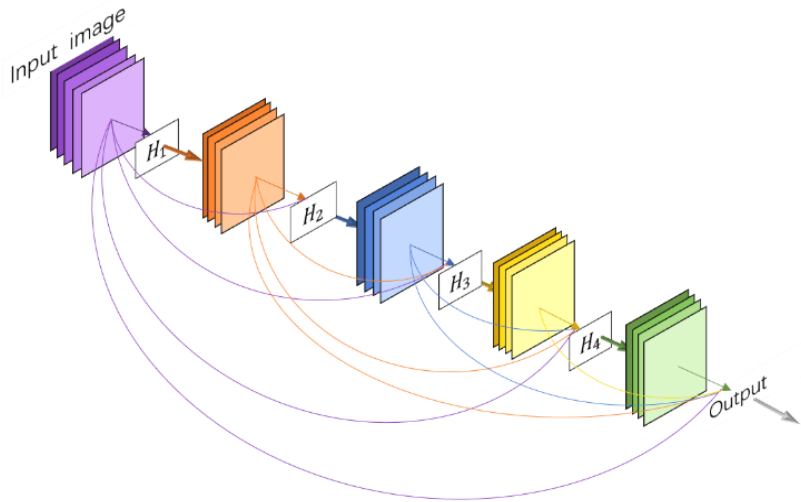

Figure 6 A 5-layer DenseNet

MobileNet proposed in [67] reduces the computational cost to 1/8~1/9 by replacing the standard convolution with a depth-separable convolution and splitting the standard convolution into one depth convolution and one point-by-point convolution. ShuffleNet proposed in [68] uses grouped convolution to reduce the number of training parameters. [69] improved ShuffleNet by dividing the input feature map into two branches and finally connecting branches and merging them with Channel Shuffle, which resulted in better identification. In 2020, [70] proposed GhostNet, which first performs a convolutional operation on the input feature map and then performs a series of simple linear operations to generate the feature map. Thus, it reduces the number of parameters and computations while achieving the performance of traditional convolutional layers. However, deep convolutional networks suffer from the loss of output spatial resolution due to the pooling layer. The mixed extraction of many different features results in less accurate information processing, such as boundaries. [71]proposed a dual-stream (shape stream and classical stream) CNN that processes information in parallel and Gated Convolutional Layer (GCL) that allows classical streams and shape streams to interact in the middle layer for better identification.

For "Head", there are usually two groups of object detection algorithms, namely two-stage object detection algorithm and one-stage object detection algorithm. The former group of algorithms first generates a series of candidate frames as samples and then classifies the samples using a convolutional neural network, which has better detection accuracy and localization precision. The latter group of algorithms does not generate candidate frames. Instead, it directly transforms the target frame localization problem into a regression problem, and the algorithm is faster. The two-stage object detection algorithms include mainly RCNN. RCNN was first proposed in [57], which first uses Selective Search to search for regions where objects may be present, then inputs these regions into AlexNet with the same size to obtain feature vectors, and finally uses SVM for classification to obtain detection results. However, RCNN has the problem of extracting features for the same region several times, so [72]proposed SPP-Net. In the feature extraction stage, it directly extracts features for a whole image to get the feature map, which avoids repeated extraction and thus improves the operation speed of the algorithm. Then, region proposals are found in the feature map, and the fixed feature vectors are extracted by spatial pyramid pooling, which greatly improves the operation speed. The Fast R-CNN proposed by [73] introduces an ROI pooling layer, the feature map output by Selective Search is ROI Pooling to get the feature vector, and softmax is used instead of SVM to classify the class of ROI compared to SPPNet. [74] proposed Faster R-CNN using Region Proposal Networks instead of Selective search. It goes through the feature map (from the convolutional layer) with a set of windows of different aspect ratios and sizes and uses these windows as candidate regions for classification and object localization. The object recognition speed and accuracy are greatly improved. However, the faster R-CNN is not shared after ROI pooling, so [75]used the full convolutional network to create shared convolution of the layers after ROI and convolve only one feature map, and the detection efficiency is further improved.

One-stage object detection algorithms include YOLO and SSD. YOLO (You Only Look Once), which treats object detection as a regression problem and uses the entire image as input to train the model. Therefore, it can detect the information of the whole picture rather than the partial picture information detected by the sliding window. Although this approach decreases accuracy, but can greatly improve the detection speed. In 2015, [76]proposed YOLOv1, as shown in Figure 7. It first splits the input image into a grid of cells and then marks the position of the object with a boundary box. If the center of the bounding box falls within a cell, this cell is responsible for predicting this object. However, when there are multiple objects whose centers fall in a single cell, the object may not be detected, or the accuracy of the detection is reduced. To solve this problem, [77] proposed YOLOv2, which removed the final fully connected layer from YOLOv1 and used convolution and anchor boxes to predict the bounding boxes. The identification rate and speed were improved. [78] proposed YOLOv3 to further improve the model by improving the single-label classification of YOLOv2 to multi-label classification and removing the pooling layer, using all convolutional layers for downsampling, and improving the network's ability to characterize the data by deepening the network and the use of FPNs. [79]proposed YOLOv4 that splits the channel into two parts, with one part performing the computation of convolution and then concatting the other part together, thus reducing the computation and being able to guarantee accuracy. [80] proposed yolov5 which is more flexible and has four network models, but the performance is not as good as YOLOv4.

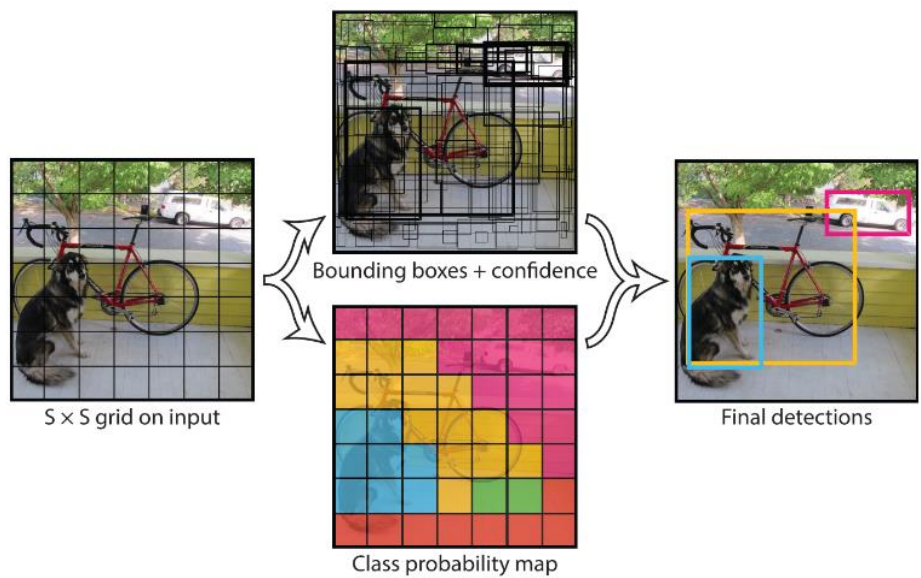

Figure 7 YOLO schematic [76] identifies the objects falling in each cell by dividing the image into a network of cells

SSD (Single Shot MultiBox Detector) was proposed by [81]. Unlike Yolo which does detection after a fully connected layer, SSD uses CNN to perform detection directly and employs a multi-scale feature map. Large-scale feature maps (near-input feature maps) can be used to detect small objects, while small-scale feature maps (near-output feature maps) are used to detect large objects, which makes SSD better than Yolo in terms of accuracy.[82] argued that there are only a few objects in many detection areas of the picture, which leads to unbalanced training sample categories and the poor performance of a one-stage object, so they proposed RetinaNet based on Cross Entropy loss. When the classification error is low, a lower weight is given. The detection performance is improved by giving higher weights for the loss when the classification error is higher.

Table 2 Comparison of object detection and identification algorithms

| Algorithm | Advantages | Disadvantages |
| --- | --- | --- |
| R-CNN | The large number of ROIs enables high detection accuracy | Training in stages, tedious steps, and high memory requirement, end-to-end training is not possible using SVM for classification |
| Fast RCNN | Using a multitasking approach to train the entire network and achieve weight sharing of network, the detection speed is improved by not repeatedly extracting features | Using the selective search extraction method takes a lot of time |
| Faster RCNN | Achieve further improvements in end-to-end training accuracy | Compared to the one-stage method, the speed is slow |
| R-FCN | Faster and slightly more accurate compared to faster R-CNN | Only one feature layer is used |
| SPP-Net | Introduce spatial pyramids to adapt to various feature maps of different sizes, with fast detection speed for one feature extraction of the whole map | The classifier cannot be trained end-to-end using SVM, and the training step is complicated because it is trained in stages |
| YOLO | Fast speed up to 45FPS, using global information of the image for prediction, low background false recognition rate | Only one object is predicted for each grid, making it easy to miss detection and difficult to detect small objects. The model training depends on labeled data, so it is not ideal for detecting uncommon objects |
| SSD | Fast speed up to 59FPS and extraction of candidate regions on feature maps of multiple sizes, with better detection results for objects of different sizes | The anchor box overlaps a lot, so it wastes lots of calculations, and it is experience-dependent and suitable for objects of fixed size |

Take Tesla as an example, the main input of the system is from cameras at eight different locations. Due to the different camera locations and poses, the image of each camera is first projected onto a virtual camera at a fixed location and pose. ResNet was then used for multi-object identification and preliminary feature extraction, and Tesla used a bi-directional feature pyramid (BiFPN) [83], which combines interlayer feature fusion with multi-resolution prediction and enables easy and fast multi-scale feature fusion. Tesla uses the Transformer self-attention mechanism to combine the camera location information and cross-learning of the features seen by each camera to integrate the images from eight cameras into a complete 360-degree, highly compressed and abstracted information. After that, introduce the timeline from the continuous video and use LSTM, which is a recurrent neural network to obtain various environmental information for auto driving.

For autonomous driving tasks, such as the detection of roads and signal light recognition, Tesla uses HydraNet [84] which has different network components for subtasks, as in Figure 8, where there is a commonly shared backbone, which splits into branches in the head. The use of feature sharing reduces repetitive convolutional computations, while the backbone network can be fixed after fine-tuning, and only the parameters of the detection head need to be trained, resulting in a significant improvement in efficiency and the ability to decouple specific tasks from the backbone.

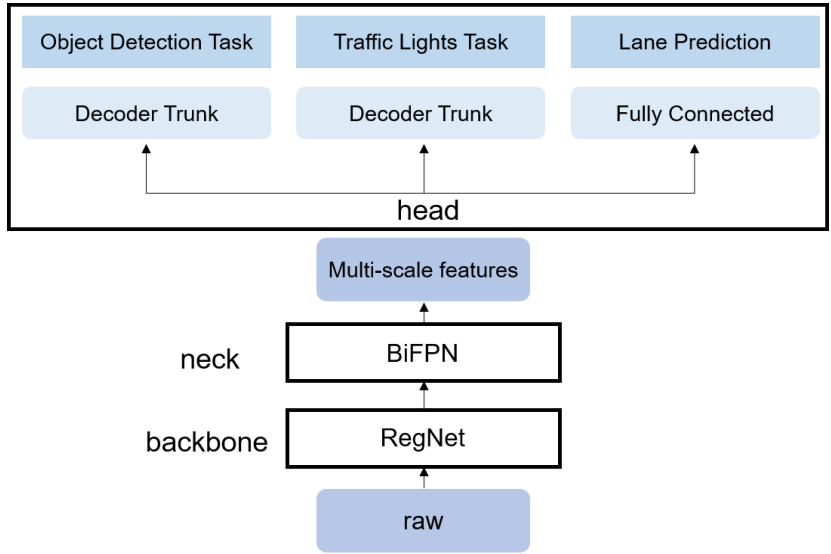

Figure 8 Multi-Task Learning HydraNets

## 2 Depth estimation

In autonomous driving systems, the proper distance is extremely important to ensure the safe driving of the car, so it requires depth estimation from the images. The goal of depth estimation is to obtain the distance to the object and finally acquire a depth map that provides depth information for a series of tasks such as 3D reconstruction, SLAM, and decision-making. The current mainstream distance measurement methods in the market are monocular, stereo, and RGBD camera-based.

### 2.1 Traditional monocular depth estimation methods

For fixed monocular cameras and objects, since the depth information cannot be measured directly, therefore, monocular depth estimation is to recognize first and then measure the distance. First, the identification is made by image matching, and then distance estimation is done based on the size of the objects in the database. Since the comparison with the established sample database is required in both the identification and estimation stages, it lacks self-learning function and the perception results are limited by the database, and the unmarked objects are generally ignored, which causes the problem that uncommon objects cannot be recognized. However, for monocular depth estimation applied to autonomous driving the objects are mainly known objects such as vehicles and pedestrians, so the geometric relationship method [85], data regression modeling method [86] and Inverse Perspective Mapping (IPM) [87] [88] can be used, and SFM(Structure From Motion)-based monocular depth estimation can be achieved by the motion of vehicles.

Currently, monocular cameras are gradually becoming the mainstream technology for visual ranging due to their low cost, fast detection speed, and ability to identify specific obstacle types, high algorithm maturity and accurate recognition.

The geometric relationship method uses the pinhole camera imaging principle. It uses light propagation along a straight line to project objects in the three-dimensional world onto a two-dimensional imaging plane, as shown in Figure 9. The vehicle distance can be calculated by the equation in the figure. However, it is required that the optical axis of the camera must be parallel to the horizontal ground, which is difficult to guarantee in practice. [89, 90] improved this by considering the yaw angle of the camera, and the distance measurement is more accurate.

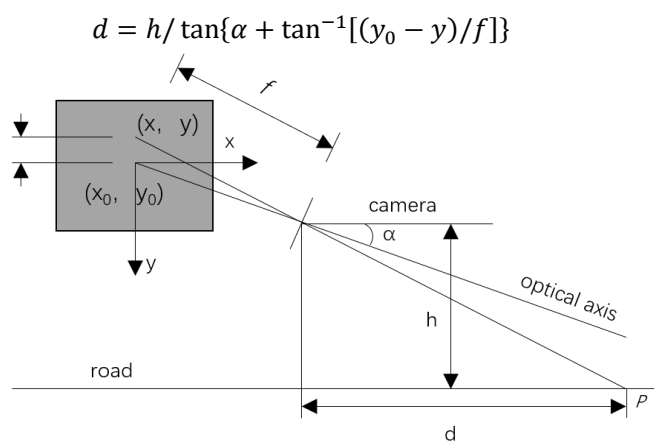

Figure 9 Geometric ranging model, α is the camera pitch angle, h is the camera height, and the projection point of the body p point in the phase plane is (x,y)

The data regression modeling method measures the distance by fitting a function to obtain a nonlinear relationship between the pixel distance and the actual distance. Inverse Perspective Mapping is widely used not only in monocular ranging but also in Around View Cameras. By converting the perspective view into a "bird's eye view," as shown in Figure 10. Since the "bird's-eye view" has a linear scale relationship with the real road plane, the actual vehicle distance can be calculated based on the pixel distance in the inverse perspective transformed view by calibrating the scale factor, which is simple and easy to implement.

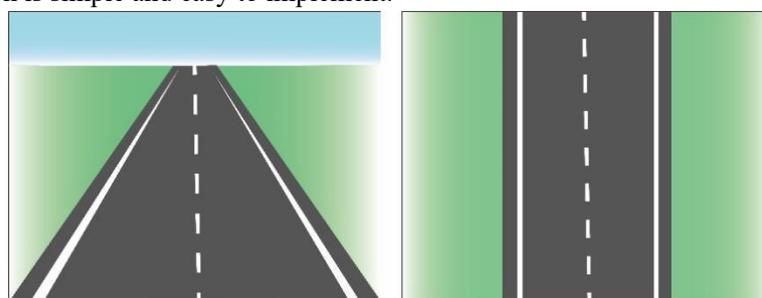

Figure 10 Transformation of the original view of the driveway into an aerial view through Inverse Perspective Mapping (right)

However, the pitch and yaw motion of the car is not considered, and the presence of pitch angle will make the inverse perspective transformed top view unable to recover the parallelism of the actual road top view, thus producing a large-ranging error. [91] proposed a distance measurement model based on variable parameter inverse perspective transformation, which dynamically compensates for the pitch angle of the camera, with a vehicle ranging error within 5% for different road environments and high robustness in real-time. However, it is impossible to calculate the pitch angle of the camera on unstructured roads without lane lines and clear road boundaries. A pitch angle estimation method without cumulative error is proposed in [92], which uses the Harris corner algorithm and the pyramid Lucas-Kanade method to detect the feature points between adjacent frames of the camera. Its camera rotation matrix and translation vector are solved by feature point matching and pairwise geometric constraints, and parameter optimization is performed using the Gauss-Newton method. Then, the pitch angle rate is decomposed from the rotation matrix, and the pitch angle is calculated from the translation vector.

SFM (Structure From Motion) is to determine the spatial geometric relationship of an object from a 2D image sequence by using mathematical theories such as multi-view geometry optimization to recover the 3D structure by camera movement. SFM is convenient and flexible but encounters scene and motion degradation problems in image sequence acquisition. According to the topology of image addition order in the process, it can be classified as incremental/sequential SFM, global SFM, hybrid SFM, and hierarchical SFM. Besides, there are semantic SFM and Deep learning-based SFM.

Table 3 Summary of major SFMs

| SFM | References | Processing | Advantages | Disadvantages |
| --- | --- | --- | --- | --- |
| incremental SFM | [93–98] | New views are added gradually based on a minimal reconstruction of two or three views, and a constraint adjustment of the scene data follows each addition | Robust as each image added requires a bundle adjustment to the data | Time consuming, and there is a danger of drift due to the accumulation of errors |
| global SFM | [99–103] | Rotation averaging is performed first to calculate the global rotation of all views and then to calculate the translation as well as the structure of each view | The global nature of the image is reflected and the risk of drift is low, time efficient | Sensitive to noise due to processing all data at once, which must be carefully filtered |
| hierarchical SFM | [104–107] | N interrelated subgraphs are obtained by dividing the original dataset, and then each subgraph is processed by incremental SFM in parallel and finally merged | Enables rapid processing of large volumes of data | The processing of each subgraph has the risk of drift and is not robust enough |

On the other hand, hybrid SFM [108] combines the advantages of incremental SFM and global SFM and is gradually becoming a trend. The pipeline can be summarized as a global estimation of a camera rotation matrix, incremental calculation of the camera center, and community-based rotation averaging method for global sensitive problems. Compared with hybrid SFM, PSFM [109] grouped the cameras into many clusters and was superior in large-scale scenes and high-precision reconstruction. [110] proposed SFMLearner to estimate the depth and pose of each frame using the photometric consistency principle. Based on this, [111] proposed SFM-Net with the addition of optical flow, scene flow, and 3D point cloud to estimate depth.

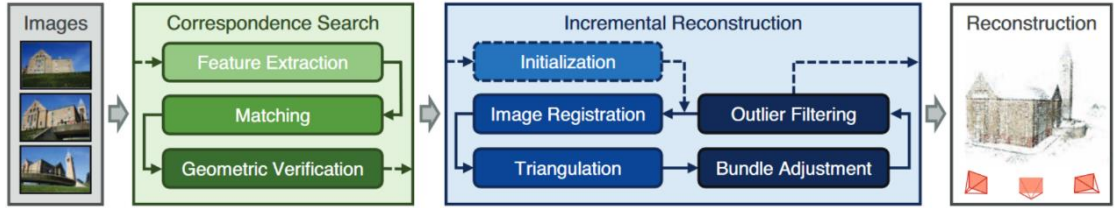

Figure 11 incremental SFM schematic diagram [112]

Monocular camera has a high proximity recognition rate, so it is widely used in front collision warning systems (FCWS). But its environmental adaptability is poor, and the camera shakes due to bumps when the vehicle is moving. In [113], a comparison experiment of three scenarios, stationary, slow-moving, and braking, resulted in taking the arithmetic mean of TTC as the alarm threshold, which can effectively circumvent abnormal situations such as camera shake and thus can be applied to more complex ranges. [114]adopted a combination of vanishing point detection, lane line extraction, and 3D spatial vehicle detection to achieve distance measurement. However, the distance error increases significantly in the case of insufficient illumination and severe obstacle occlusion in front. In [115], it is proposed that the absolute scale and attitude of the system are estimated using monocular Visual Odometry combined with the GPS road surface characteristics and geometrics prior to detecting and ranging the object in front of the vehicle and that the localization of the camera itself and the object can be achieved using the 3D shape change of the object.

**2.2 Deep Learning-based monocular depth estimation**

The input of Deep Learning-based monocular depth estimation is the captured original image, and the output is a depth map on which every pixel value corresponds to the scene depth of the input image. Deep Learning-based monocular depth estimation algorithms are divided into supervised and unsupervised learning. Supervised learning is able to recover scale information from the structure of individual images and scenes with high accuracy because they train the network directly with ground truth depth values but require datasets such as KITTI, Open Image, Kinetics, JFT-300M, etc. Since depth data is difficult to obtain, a large number of algorithms are currently based on unsupervised models.

[116] used Markov Random Field (MRF) to learn the mapping relationship between input image features and output depth, but the relationship between RGB images and depth needs to be artificially assumed. The model is difficult to simulate the real-world mapping relationship, and therefore the prediction accuracy is limited. In 2014, [117] proposed to convolve and downsample the image in multiple layers to obtain descriptive features of the whole scene and used them to predict the global depth. Then, the local information of the predicted image is refined by a second branching network, where the global depth will be used as input to the local branch to assist in the prediction of the local depth. In 2015, [118] proposed a unified multi-scale network framework based on the above work. The framework uses a deeper base network, VGG, and uses a 3rd fine-scale network to add detailed information further to improve the resolution for better depth estimation. In 2016, [119] used a convolutional neural network model for vehicle detection and localization and then calculated the distance based on the monocular vision principle. In 2018, [120]

proposed the DORN framework to divide continuous depth values into discrete intervals and then used fully connected layers to decode and inflate the convolution for feature extraction and distance measurement. In the same year, [121] compared LiDAR to convert the input image into point cloud data similar to that generated by LiDAR and then used point cloud and image fusion algorithms to detect and measure distance. [122] proposed MonoGRNet. It gets the visual features of the object by ROIAlign and then uses these features to predict the depth of the 3D center of the object. In 2019, [123] improved it by proposing MonoGRNetV2 to extend the centROIds to multiple key points and use 3D CAD object models for depth estimation. [124] proposed BEV-IPM to convert the image from a perspective view to a bird's eye view circle (BEV). In the BEV view, the Bottom Box (the contact part between the object and the road surface) is detected based on the YOLO network. Then its distance is accurately estimated using the Box predicted by the neural network. [125]suggested the use of multi-scale feature maps output by convolutional neural networks to predict the depth maps of different resolutions based on the depth estimation of two resolutions, and the feature maps of different resolutions are fused by successive MRFs to obtain the depth maps corresponding to the input images. [126] proposed 3D-RCNN, which first uses PCA to downscale the parameter space and then generates 2D images and depth maps based on each target low-dimensional model parameter predicted by the R-CNN. Nevertheless, CNNs can handle global information better only when at lower spatial resolutions. The key to the effectiveness of monocular depth estimation enhancement is that sufficient global analysis should be performed on the output values. Therefore, 2020 [127]proposed the AdaBins structure, which combines CNN and transformer. Using the excellent global information processing capability of the transformer, combined with the local feature processing capability of CNN, the accuracy of depth estimation is greatly improved.

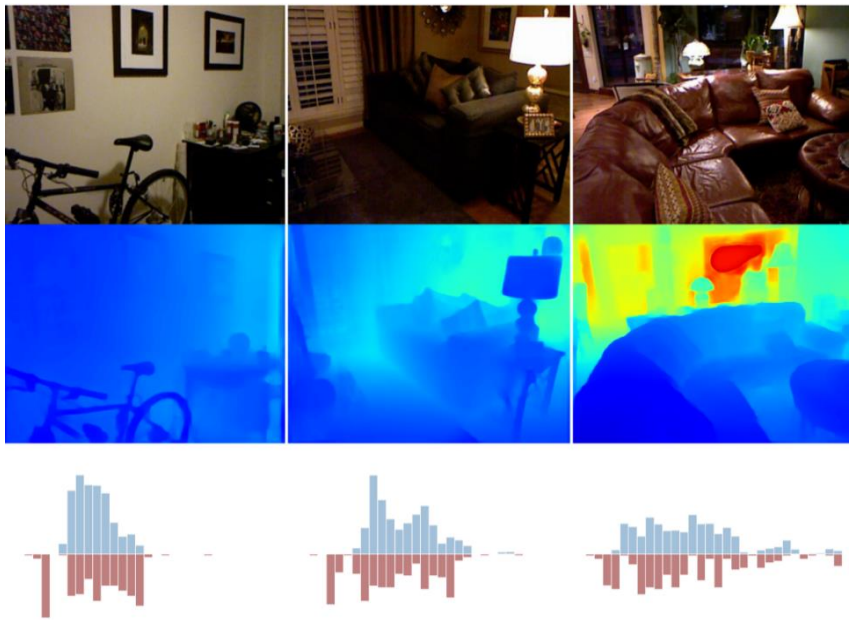

Figure 12 AdaBins [127]top is original. The middle is the depth map, and the bottom is the histogram of depth values of the ground truth and predicted adaptive depth

According to [128], an end-to-end convolutional neural network framework is utilized for vehicle ranging in order to cope with measurement errors due to light variations and viewpoint variations. The algorithm is based on converting RGB information into depth information, combined with a detection module as input, and finally predicting the distance based on the distance module. Its robustness is better and reduces the ranging error due to the complex driving environment, such as insufficient light and occlusion. In 2021, [129]proposed FIERY, an end-to-end BEV probabilistic prediction model, which inputs the current state captured by the camera and the future distribution in training to a convolutional GRU network for inference as a way to estimate depth information and predict future multimodal trajectories.

Table 4 Summary of monocular depth estimation

| Method | References | Principle | Features | With or without supervision |
|---|---|---|---|---|
| Classification-based approach | [120][130, 131] | Divide the depth values into many classes and determine the pixel depth values by training CNN | When classifying, the model assigns all predicted depths in a range to the same interval median so accuracy is low | Supervised |
| Regression-based methods | [132–134] | Modele the hidden mapping function between RGB images and the corresponding mappings | Dataset dependent, complex network structure, slow computation speed, and easy to fall into local optimal solutions | Supervised |
| Conditional random field-based approach | [135–137] | A conditional random field describes the relationship between the depth of a pixel and its neighboring pixels to find the depth value that matches the real scene | Adding CRF after CNN can improve performance, do not depend on geometric priors, and enable end-to-end learning | Supervised |
| Stereo image-based methods | [138–141] | Using binocular images as input and labels to train the network to predict monocular parallax and then calculate the depth information | The prediction results are biased due to occlusion and boundary area problems | Unsupervised |
| Video sequence-based approach | [111, 142, 143] | Multiple continuous images are used as input to perform depth estimation using photometric loss | It reduces the usage limit of the data and improves the estimation performance by training with multiple frames | Unsupervised |

**2.3 Traditional stereo depth estimation methods**

Unlike the monocular camera, stereo depth estimation relies on the parallax produced by cameras arranged in parallel. It can obtain depth information about the driveable area and obstacles in the scene by finding points of the same object and making accurate triangulation. Despite not being as far as the LIDAR depth estimation, it is cheaper and can reconstruct 3D information of the environment when there is a common field of view. However, stereo cameras require a high synchronization rate and sampling rate between cameras, so the technical difficulty lies in stereo calibration and stereo positioning. Among them, the binocular camera is the most used as shown in Figure 13.

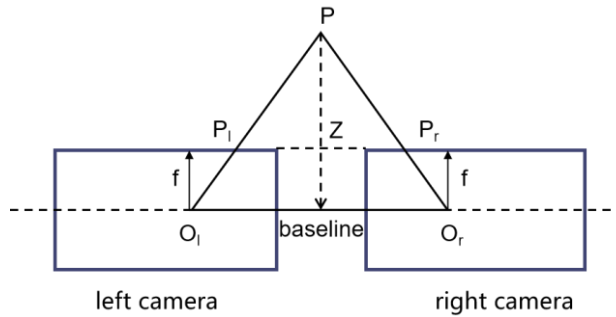

Figure 13 Binocular distance measurement schematic

The working principle of the trinocular camera is equivalent to using two binocular stereo-vision systems. They are placed along the same direction and at the same distance, as shown in Figure 14. The trinocular stereo-vision system has a narrow baseline and a wide baseline. The narrow baseline is the line of the left and middle cameras and the wide baseline is the line of the left and right cameras. The narrow baseline increases the field of view common to both cameras, and the wide baseline has a larger maximum field of view at each visible distance [144]. The three cameras of the trinocular stereo vision system take three terrain images from different angles and then use the stereo vision matching algorithm to obtain the depth information of the terrain.

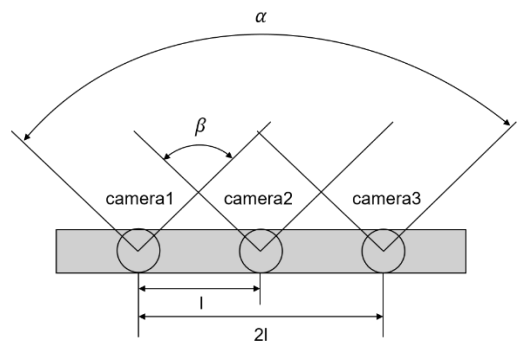

Figure 14: Schematic diagram of the trinocular camera, where α is the maximum field of view of the left and right cameras, and β is the common field of view of the left-center camera

Similar to monocular, stereo ranging works on the principle of affine transformation of the actual object as it is captured by the camera into the picture. The process includes calibration of the camera, stereo correction of the image, calculation of the parallax map, and calculation of the depth map. Due to parallax, the stereo vision system requires stereo matching of the corresponding points captured in different images. Stereo matching is mainly divided into global matching and local matching. Although global matching has high accuracy and better robustness, the computation speed is slow and cannot meet the real-time requirements, so local matching is mainly applied for vehicles.

In [145] [146], the vehicle distance measurement is achieved using the center point coordinates based on matching the feature points of the vehicle marker and the license plate, respectively. [147] improved the matching speed and accuracy by extracting the Harris corner points of 3D images as feature points and using the wavelet transform sublinear matching method. Still, it does not have scale invariance and is only applicable to close-range vehicle distance measurement due to the weak interference immunity of Harris. [148] proposed a binocular stereo vision calibration method based on parallel optical axes. Its binocular visual obstacle detection matching is achieved by matching left and right images using the SIFT algorithm. The SIFT feature points can maintain good invariance when the scale size of the image changes or when the image is rotated or panned and still maintain certain stability when the light intensity and the angle of the camera change, thus avoiding certain noise interference. [149] is positioned by combining with high-precision maps. The current road information is obtained by stereo vision camera, and preliminary mapping of position is performed by using Kalman filter to match with map road marking information. By using the ORB algorithm (Oriented FAST and Rotated BRIEF) for feature point matching left and right eye and front and back frame to extract the consistency information in the image sequence, the position change of consistency information in the image sequence is selected for camera motion estimation. Thus the camera pose is derived from achieving ranging. However, due to the large number and uneven distribution of feature points extracted by the traditional ORB algorithm, the accuracy of this stereo-depth estimation method is low. In [150], a vehicle distance measurement method based on machine learning and an improved ORB algorithm is proposed. The method uses dynamic thresholding to improve the quality of feature point extraction and the Progressive Sample Consensus (PROSAC) algorithm in the feature matching stage to reduce mismatching to improve the front vehicle distance measurement accuracy. [151] proposed the concept of SurroundDepth. The depth map between cameras is obtained by processing multiple surrounding views and fusing the information from multiple views with a cross-view transformer.

Table 5 Comparison of stereo matching algorithms

| Algorithm | Advantages | Disadvantages |
| --- | --- | --- |
| Harris | Simple calculation, uniform extracted point features, insensitive to image rotation, brightness change, noise effect, and viewpoint transformation | No scale invariance and the extracted corner points are pixel level |
| SIFT | Good stability and invariance, it can adapt to rotation, scale scaling, and brightness changes, and can be free from the interference of perspective changes, affine transformations, and noise | Only the grayscale nature algorithm is utilized, neglecting the color information, running slowly, and detecting too few feature points for blurred images and images with smooth edges |
| SURF | The improved version of SIFT is more than three times faster in computation and more representative in the extraction of feature points | Not real-time, too dependent on the gradient direction of the pixels in the local area in the main direction-finding stage, may not be accurate in the main direction |
| ORB | Good real-time performance, the calculation speed is only 1% of SIFT, saving storage space | Rotation and fuzzy robustness is poor and not scale invariant |

**2.4 Deep Learning-based stereo depth estimation**

Traditional stereo-based depth estimation is achieved by matching the features of multiple images. Despite extensive research, it still suffers from poor accuracy when dealing with occlusion, featureless regions, or highly textured regions with repeating patterns. In recent years, stereo depth estimation based on Deep Learning has developed rapidly, and the robustness of depth estimation has been greatly improved by using prior knowledge to characterize features as a learning task.

In 2016, [152] proposed MC-CNN to construct a training set by labeling data. A positive sample and a negative sample are generated at each pixel point, where the positive samples are from two image blocks with the same depth, and the negative samples are from image blocks with different depths, and then the neural network is trained to predict the depth. However, its computation relies on local image blocks, which introduces large errors in some regions with less texture or recurrence of patterns. Therefore, in 2017 [153] proposed GC-Net, which performs multilayer convolution and downsampling operations on the left and right images to extract semantic features better, and then uses 3D convolution to process Cost Volumn so as to extract correlation information between the left and right images as well as parallax values. In 2018, [154] proposed PSMNet, which employs pyramidal structures and null convolution to extract multifractional aspect information and expand the field of perception and multiple stacked HourGlass structures to enhance 3D convolution so that the estimation of parallax relies more on the information at different scales rather than local information at the pixel level. Thus, a more reliable estimate of parallax can be obtained. [155] proposed MVSNet, which utilizes 3D convolutional operation cost volume regularization. It first outputs the probability of each depth. It then finds the weighted average of the depths to obtain the predicted depth information, using reconstruction constraints (photometric and geometric consistencies) between multiple images to select the predicted correct depth information. In 2019, [156] proposed P-MVSNet based on it, which makes a better estimation structure by a hybrid 3D Unet with isotropic and anisotropic 3D convolution. However, these networks use discrete points for depth estimation, thus introducing errors. [157] considering that existing stereo networks (e.g., PSMNet) produce parallax maps that are not geometrically consistent, they propose StereoDRNet, which takes as input geometric errors, photometric errors, and undetermined parallax, to produce depth information and predict the occluded portion. This approach gives better results and significantly reduces computational time. In 2020, [158]proposed a CDN for continuous depth estimation. In addition to the distribution of discrete points, the offset at each point is also estimated, and the discrete points and the offset together form a continuous parallax estimate.

**2.5 RGBD distance measurement**

RGBD camera generally contains three cameras: a color camera, an IR transmitter camera, and an IR receiver camera, and the principle is shown in Figure 15. Compared with stereo cameras, which calculate depth by parallax, RGB-D can actively measure the depth of each pixel. Moreover, the 3D reconstruction based on RGBD sensors is cost-effective and accurate, which makes up for the computational complexity of monocular and stereo vision sensors to estimate depth information and the lack of guaranteed accuracy.

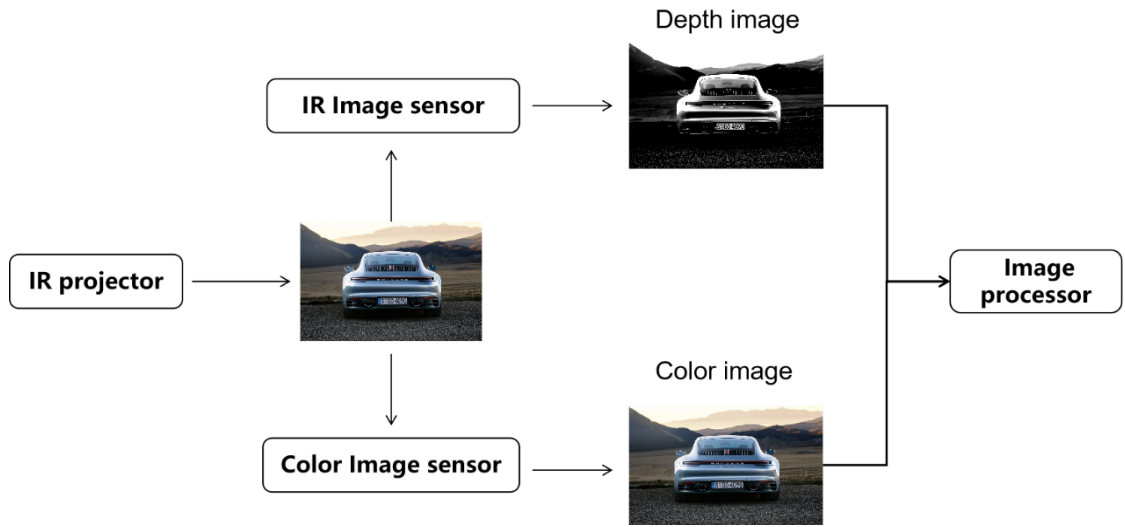

Figure 15 RGB-D schematic

RGB-D measures the pixel distance, which can be divided into the infrared structured light method and the time-of-flight (TOF) method. The principle of structured light [159] is that an infrared laser emits some patterns with structural features to the surface of an object. Then an infrared camera will collect the pattern variations due to the different depths of the surface. Unlike stereo vision which relies on the feature points of the object itself, the structured light method characterizes the transmitted light source, therefore, the feature points do not change with the scene, which greatly reduces the matching difficulty. According to the different coding strategies, there are temporal coding, spatial coding, and direct coding. The temporal coding method can be classified into binary codes [160], n-value codes [161], etc. It has the advantages of easy implementation, high spatial resolution, and high accuracy of 3D measurement, but the measurement process needs to project multiple patterns, so it is only suitable for static scene measurement. Spatial coding method has only one projection pattern, and the code word of each point in the pattern is obtained based on the information of its surrounding neighboring points (e.g., pixel value, color, or geometry). It is suitable for dynamic scene 3D information acquisition, but the loss of spatial neighbor point information in the decoding stage leads to errors and low resolution. Spatial coding is classified into informal coding [162], De Bruijn sequence-based coding [163], and M-array-based coding [164]. Direct coding method is performed for each pixel. However, it has a small color difference between neighboring pixels, which will be quite sensitive to noise. It is unsuitable for dynamic scenes, including the gray direct coding proposed by [165] and the color direct coding proposed by [166].

TOF calculates the distance of the measured object from the camera by continuously emitting light pulses to the observed object and then receiving the light pulses reflected back from the object and by detecting the time of flight of the light pulses. Depending on the modulation method, it can be generally divided into Pulsed Modulation and Continuous Wave Modulation.

After measuring the depth, RGB-D completes the pairing between depth and color pixels according to the individual camera placement at the time of production and outputs a one-to-one color map and a depth map. The color information and distance information can be read at the same image location, and the 3D camera coordinates of the pixels can be calculated to generate a point cloud. However, RGB-D is susceptible to interference from daylight or infrared light emitted by other sensors, so it cannot be used outdoors. Multiple RGB-Ds can interfere with each other and have some disadvantages in terms of cost and power consumption.

# 3 Vision SLAM

SLAM (Simultaneous Location and Mapping) is divided into laser SLAM and visual SLAM. Laser SLAM has some disadvantages, such as lack of color information, high price, and insufficient effective distance. Vision SLAM uses a vision sensor as the only environmental perception sensor. The triangulation algorithm of a single vision sensor or the stereo matching algorithm of multiple vision sensors can calculate the depth information with good accuracy. At the same time, because it contains rich color and texture information and has the advantages of small size, lightweight, and low cost, it has become a current trend of research. Vision SLAM is divided into monocular vision SLAM, stereo vision SLAM, and RGB-D vision SLAM, depending on the vision sensor category.

**3.1 Monocular vision SLAM**

The monocular SLAM is a simple, low cost and easy-to-implement system using a camera as the only external sensor. Monocular vision SLAM is divided into two types according to whether a probabilistic framework is used or not. Monocular vision SLAM based on a probabilistic framework constructs a joint posterior probability density function to describe the spatial locations of camera poses and map features given the control inputs from the initial moment to the current moment and the observed data, and then estimates this probability density function by a recursive Bayesian filtering method, which is currently widely used because of the unknown complexity of SLAM application scenarios.[167]proposed an example filter-based SLAM that decomposes the joint posterior distribution estimation problem of motion paths and maps it into the estimation problem of motion paths with example filters and the estimation problem of road signs under known paths. However, to ensure localization accuracy, more particles are required in complex scenes and motions, which greatly increases the complexity of the algorithm, and resampling tends to lead to sample depletion and other problems. [168] improved the particle filtering method by marginalizing the position of each particle feature to obtain the probability that the sequence of observations of that feature is used to update the weights of the particles and does not require the feature positions included in the state vector. Consequently, the algorithm's computational complexity and sample complexity remain low even in feature-dense environments.[169] proposed MonoSLAM based on Extended Kalman Filtering, which uses sparse feature maps to represent the environment and actively search for and track feature points through probabilistic estimation. However, the EKF-SLAM algorithm suffers from high complexity, poor data association problems, and large linearization processing error. [170] proposed FastSLAM, which still uses EKF algorithm to estimate the environmental features, but the computational complexity is greatly reduced by representing the mobile robot's poses as particles and by decomposing the state estimation into a sampling part and a resolution part. However, its use of the process model of SLAM as a direct importance function for the sampled particles may lead to the problem of particle degradation, which reduces the accuracy of the algorithm. Therefore, FastSLAM2.0 proposed in [171] uses EKF algorithm to recursively estimate the mobile robot poses, obtain the estimated mean and variance, and use them to construct a Gaussian distribution function as the importance function. Thus, the particle degradation problem is solved. For monocular vision SLAM with the non-probabilistic framework, [172] proposed a keyframe-based monocular vision SLAM system PTAM. The system utilizes one thread to track the camera pose and another thread to bundle and adjust the keyframe data as well as the spatial positions of all feature points. The dual-threaded parallelism ensures both the accuracy of the algorithm and the efficiency of the computation.[173] proposed ORB-SLAM based on PTAM with the addition of a third parallel thread, the loopback detection thread, and the loopback detection algorithm can reduce the cumulative error generated by the SLAM system. And because of the rotation and scale invariance of ORB features, the endogenous consistency and good robustness in each step are ensured. A comparison of the two is shown in Figure 16. [174, 175] proposed ORB-SLAM2 and ORB-SLAM3 and extended them to binoculars, RGBD, and fisheye cameras. [176] utilized a fixed number of images recently captured by the camera as key frames for local bundle adjustment optimization to achieve SLAM. [177] presents LIFT-SLAM, which combines deep learning-based feature descriptors with a traditional geometry-based system. Features are extracted from images using CNNs, which provide more accurate matching based on the features obtained from learning.

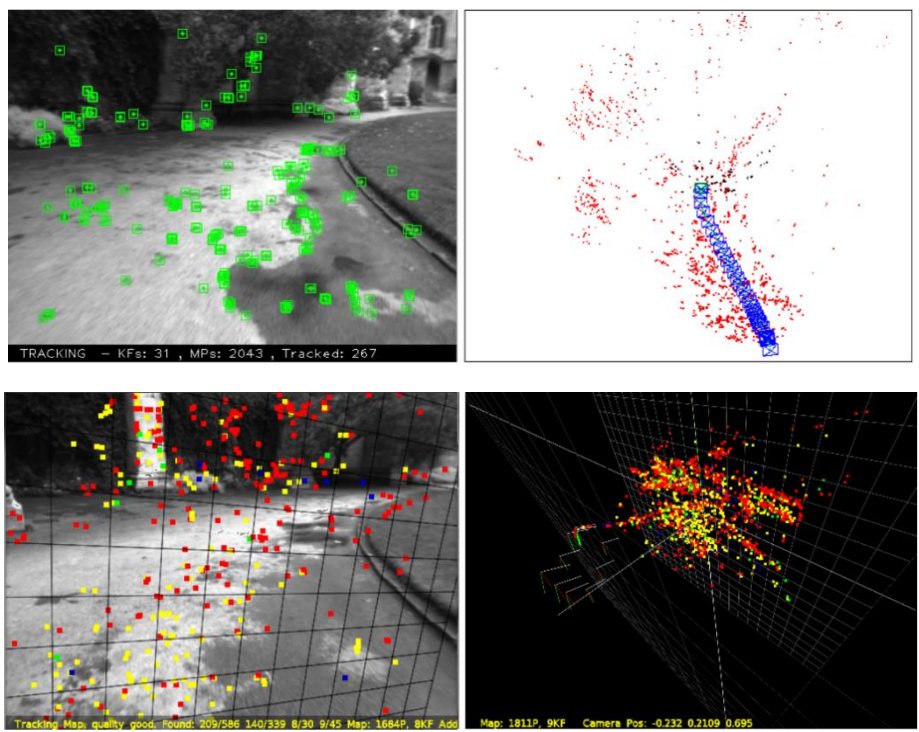

Figure 16 Compared with ORB-slam (top) and PTAM (bottom) initialization maps, the method of the fundamental matrix can explain more complex scenes. [173]

## 3.2 Stereo vision SLAM

Stereo vision SLAM uses multiple cameras as sensors. Since the absolute depth is unknown, monocular SLAM cannot get the true size of the motion trajectory and map. Stereo can calculate the true 3D coordinates of the landmarks in the scene simply and accurately through the parallax. However, it requires higher accuracy of calibration parameters and is costly. In [178], the proposed LSD-SLAM by fixing the baseline can avoid the scale drift that usually occurs in monocular SLAM, and in addition, by combining two parallax sources can estimate the depth of the under-constrained pixels. ORB-SLAM2 applied to stereo vision SLAM uses dual threads to extract ORB feature points for the left and right images, and then computes binocular seeing feature points and performs matching. [179] proposes DMS-SLAM based on ORB-SLAM 2.0 using sliding window and grid-based motion statistics (GMS) feature matching methods to find static feature locations with some improvement in execution speed. However, point feature-based algorithms do not work well in low-texture environments, so [180]proposes PL-SLAM combining point and line features based on ORB-SLAM2 and LSD, which can guarantee robust performance in a wider range of scenarios. [181] proposed a framework for stereo vision Dual Quaternion Visual SLAM. It uses a Bayesian framework for pose estimation, and for point clouds and optical flow of the map, DQV-SLAM uses an ORB function to achieve reliable data correlation in a dynamic environment. The performance is better compared to filter-based methods. [182]proposed SOFT, a stereo vision odometry computation based on feature tracking. SLAM is achieved by bit pose estimation and construction of feature point-based bit pose maps, and the global consistency of the system is guaranteed compared to monocular ORB. However, the binocular camera degrades to a monocular when the target is far away. Therefore, a large amount of research has been conducted around monocular ORB in recent years.

**3.3 RGB-D vision SLAM**

RGB-D vision SLAM uses an RGB-D sensor as the image input device. This sensor integrates a color camera and an infrared camera to capture both color images and the corresponding depth images and thus is gradually becoming the trend of SLAM. [183] extracted feature points from RGB images and then combined them with depth information to inverse map the feature points to 3D space and then optimized the initial poses using ICP which is a point cloud matching algorithm. However, the RGB-D data often lacks validity when in an environment with changing light intensity, so there is now a fusion of it with the state increments calculated by the IMU sensor [184] to obtain better results. The current deep SLAM consists of two parts, the front end, and the back end. The front-end estimates camera motion from images between adjacent frames and recovers the spatial structure of the scene, while the back-end receives the camera pose output from the visual odometry at different moments, as well as information from loopback detection, and optimizes them to obtain globally consistent trajectories and maps. [185] used Kinect RGB-D for 3D environment reconstruction. KinectFusion technology can add every frame of acquired image data to the 3D map in real-time. Still, it requires high hardware configuration because it occupies huge memory space, and the performance of SLAM deteriorates for a long time.[186] proposed improvement and optimization of the RGB-D SLAM problem based on it. The front-end of the system extracts features from the RGB image of each frame, combines both RANSAC and ICP algorithms to obtain motion estimates, and uses EMM (Environment Measurement Model) model to validate the motion estimates. The back end builds a map based on pose graph optimization.

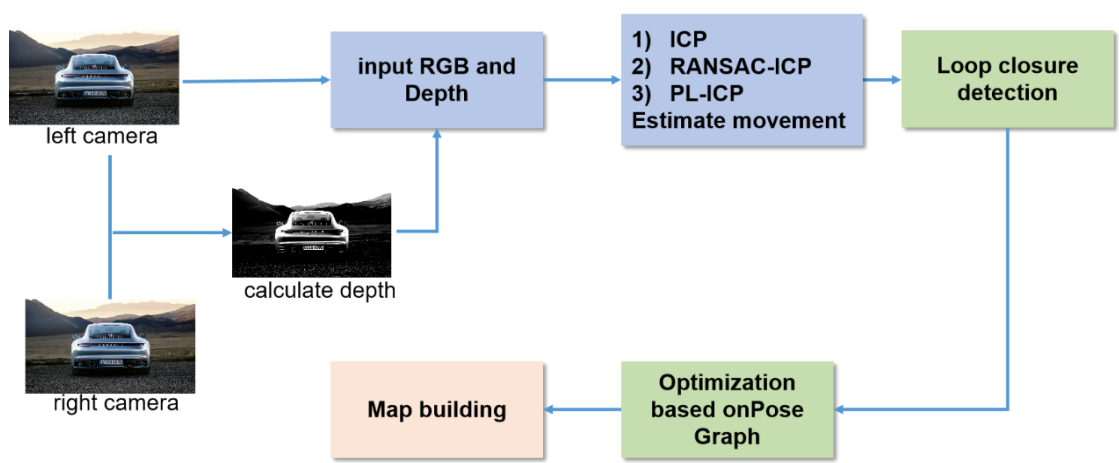

Figure 17 RGBD SLAM flow chart, depth map obtained from left and right camera images, then motion estimation and scene building by various algorithms at the front end, loopback detection and optimization at the back end

## 4 Conclusion

With the rapid development of autonomous driving technology, environmental perception is the basis and prerequisite for vehicles to realize autonomous driving. Through the acquisition, processing, and analysis of environmental images, it finally gets the information about the road, pedestrians, and obstacles around the vehicle and then makes decisions and controls.

In this paper, the application of vision technology in the field of autonomous driving is described in a systematic way. In the field of object detection and identification, traditional methods first preprocess the acquired images, including image compression, image enhancement and recovery, and image segmentation. Then, the edge features, appearance features, statistical features, transformation coefficient features, and other features of the images are compared with existing patterns to determine their categories. Image detection and identification technology combined with Deep Learning needs more time in training. However, its reliability and processing speed are better, so it has become the future trend of object identification, and this paper introduces and compares the networks and algorithms for image processing. Currently, the mainstream recognition algorithms include the RCNN series, YOLO series, FCN, and SSD, etc. Faster R-CNN has higher accuracy mAP and lower miss detection rate but slower speed. On the contrary, SSD balances the advantages and disadvantages of YOLO and Faster RCNN with better accuracy and speed. For the problem of blind spot detection, the Around View Camera can stitch together the images from all directions on top of the car to effectively avoid blind spots and greatly reduce traffic accidents. Its key technologies include image correction, top view transformation, and image stitching.

In depth estimation, the monocular camera dominates a large market with its advantages of low cost and fast detection. Its measurement methods mainly include the geometric relationship method, data regression modeling method, inverse perspective transformation method, SFM based method and so on. This paper also compares the principles and advantages, and disadvantages of different SFMs, among which hybrid SFM combines the advantages of incremental and global styles with better results. For Deep Learning based monocular depth estimation, this paper introduces and compares some algorithms, including supervised methods based on classification, regression, conditional random fields, and unsupervised methods based on stereo images and video sequences. Stereo cameras directly perform distance measurements through parallax, although they do not rely on sample datasets. However, camera calibration and fusion are required, and the commonly used algorithms are SIFT, SUFT, and ORB. As for monocular, it can obtain the complete 3D information of the road for the recognition of road information, thus providing a reference for decision-making modules such as obstacle avoidance and path planning. In contrast, since RGB-D does not depend on the fixed characteristics of the object itself, it can actively measure the depth of each pixel with high accuracy through infrared transmitting and receiving devices.

Visual SLAM has become a current research trend with its advantages of rich color and texture information, small size, lightweight, low cost, and real-time. The monocular visual SLAM has a simple structure. Among them, the probabilistic framework-based approach describes the spatial location of camera poses and map features by constructing a joint posterior probability density function, which is better adapted for unknown environments. The stereo vision SLAM can simply and accurately calculate the true 3D coordinates of the landmark points in the scene through the view source difference and can estimate the depth of the under-constrained pixels. RGB-D vision SLAM consists of image processing front-end and back-end optimization, where the front-end extracts features from each RGB image frame and performs motion estimation. The back end optimizes and builds real-time maps based on pose graphs.

## 5 Future Outlook

A. Visual perception performance

Although we have advanced vision sensor technologies, there is still the problem of decreasing accuracy in bad weather. [187] experimentally concluded that the detection error reaches 6%-17% in rain and snow, and this error lays a hidden danger for traffic accidents. For depth estimation, binocular images need to use stereo matching for pixel point correspondence and parallax calculation, so the computational complexity is also higher, especially for low-texture scenes where matching is not effective. Monocular depth image processing relies heavily on large-scale labeled datasets and is less effective for unknown unlabeled target recognition. Hence, unsupervised learning techniques are still a future research direction. Furthermore, road driving will also affect the recognition and range of objects in multi-vehicle and inter-vehicle occlusion situations. Thus long-range distance measurement in complex scenes will be a hot spot that still needs to be researched in the future.

B. High speed and network lightweighting

For autonomous driving, real-time is very important. Therefore, fast image segmentation is needed for the purpose of fast Region of Interest (ROI) recognition. In the future development, the processing speed needs to be further improved. Currently, methods for identifying distance measurements usually rely on highly parameterized deep neural networks to achieve high accuracy. This implies the huge number of parameters and slow inference speed of the model, so lightweight networks like SqueezeNet [188] are also a hot topic for the future.

C. Simultaneous Location and Mapping

In the past decades, visual SLAM has developed rapidly and is used in the field of autonomous driving, and is very good at perception and localization tasks. However, the current visual SLAM still suffers from the problems of difficulty in balancing real-time and accuracy, sensitivity to light, and low robustness in poor lighting conditions or complex lighting situations. At the same time, the application in outdoor dynamic and complex scenes still faces great challenges. Moreover, the combination of SLAM and Deep Learning is also a future development trend. Using neural networks can better extract semantic information and thus improve the generalization and reconstruction performance of the model.